\documentclass[letterpaper]{article} 
\usepackage{aaai24}  
\usepackage{times}  
\usepackage{helvet}  
\usepackage{courier}  
\usepackage[hyphens]{url}  
\usepackage{graphicx} 
\usepackage{natbib}  
\usepackage{caption} 
\usepackage{algorithm}
\usepackage{algorithmic}
\usepackage{graphicx}
\usepackage{amsmath}
\usepackage{amssymb}
\usepackage{booktabs}
\usepackage{multirow} 
\usepackage{times}
\usepackage{epsfig}
\usepackage{newfloat}
\usepackage{listings}
\DeclareCaptionStyle{ruled}{labelfont=normalfont,labelsep=colon,strut=off} 
\floatstyle{ruled}
\newfloat{listing}{tb}{lst}{}
\floatname{listing}{Listing}
\title{Attention Disturbance and Dual-Path Constraint Network \\ for Occluded Person Re-identification}
\author {
    Jiaer Xia\textsuperscript{\rm 1}\thanks{Equal contribution.},
    Lei Tan\textsuperscript{\rm 1}\footnotemark[1],
    Pingyang Dai\textsuperscript{\rm 1}\thanks{Corresponding author.},
    Mingbo Zhao\textsuperscript{\rm 2},
    Yongjian Wu\textsuperscript{\rm 3},
    Liujuan Cao\textsuperscript{\rm 1}
}

\affiliations {
    \textsuperscript{\rm 1}Key Laboratory of Multimedia Trusted Perception and Efficient Computing,\\
Ministry of Education of China, Xiamen University, China\\
    \textsuperscript{\rm 2}Donghua University, Shanghai, China\\
    \textsuperscript{\rm 3}Tencent Youtu Lab, Shanghai, China\\
    \{xiajiaer, tanlei\}@stu.xmu.edu.cn, pydai@xmu.edu.cn,\\
    mzhao4@dhu.edu.cn, littlekenwu@tencent.com, caoliujuan@xmu.edu.cn
}

\usepackage{bibentry}

\begin{document}

\maketitle

\begin{abstract}
Occluded person re-identification (Re-ID) aims to address the potential occlusion problem when matching occluded or holistic pedestrians from different camera views. Many methods use the background as artificial occlusion and rely on attention networks to exclude noisy interference. However, the significant discrepancy between simple background occlusion and realistic occlusion can negatively impact the generalization of the network.
To address this issue, we propose a novel transformer-based Attention Disturbance and Dual-Path Constraint Network (ADP) to enhance the generalization of attention networks. Firstly, to imitate real-world obstacles, we introduce an Attention Disturbance Mask (ADM) module that generates an offensive noise, which can distract attention like a realistic occluder, as a more complex form of occlusion.
Secondly, to fully exploit these complex occluded images, we develop a Dual-Path Constraint Module (DPC) that can obtain preferable supervision information from holistic images through dual-path interaction. With our proposed method, the network can effectively circumvent a wide variety of occlusions using the basic ViT baseline. Comprehensive experimental evaluations conducted on person re-ID benchmarks demonstrate the superiority of ADP over state-of-the-art methods.
\end{abstract}


\section{Introduction}
\label{sec:intro}
%
Person re-identification (Re-ID) refers to the process of matching pedestrian images captured by non-overlapping cameras. This technique has gained popularity in recent years as surveillance systems have become more advanced and widespread.
%
With the rapid development of deep learning technology~\cite{DBLP:conf/cvpr/HeZRS16, DBLP:conf/nips/VaswaniSPUJGKP17, DBLP:conf/iclr/DosovitskiyB0WZ21}, Re-ID has also achieved remarkable performance~\cite{DBLP:conf/cvpr/0004GLL019,DBLP:conf/cvpr/ZhaiLYSCJ020,DBLP:conf/iccv/ZhengSTWWT15,DBLP:conf/nips/EomH19,DBLP:journals/pami/ChenZZL18,DBLP:conf/wacv/WuCLWYZ16} by meriting from its powerful feature extraction capabilities.
%
%
However, most existing methods assume that the pedestrians in retrieved images are unobstructed, ignoring the possible occlusion problems that can occur in real-world scenarios.
%
%
Consequently, these methods significantly degrade when dealing with occluded images. 
%
%
While recent endeavors have facilitated person Re-ID under occlusion conditions~\cite{DBLP:conf/iccv/YanPJ0FS21, DBLP:conf/cvpr/WangZT0HS22, DBLP:conf/mm/TanDJW22, DBLP:conf/cvpr/LiHZL0021, DBLP:journals/ijon/ShiLWZL22,jia2022learning}, two main problems associated with occlusions still need to be addressed.
Firstly, the presence of obstacles will vanish some parts of the human body, missing and misaligned extracted features.
%
%
Traditional Re-ID methods cannot perform valid retrievals when some discriminative parts are obscured.
%
%
Secondly, occlusions introduce noise into extracted features, polluting the final feature representation of each image. When dealing with these polluted features, different identities may have high similarities due to the same obstacle, resulting in incorrect matches.
\begin{figure}[t]
  \centering
   \includegraphics[width=1.0\linewidth]{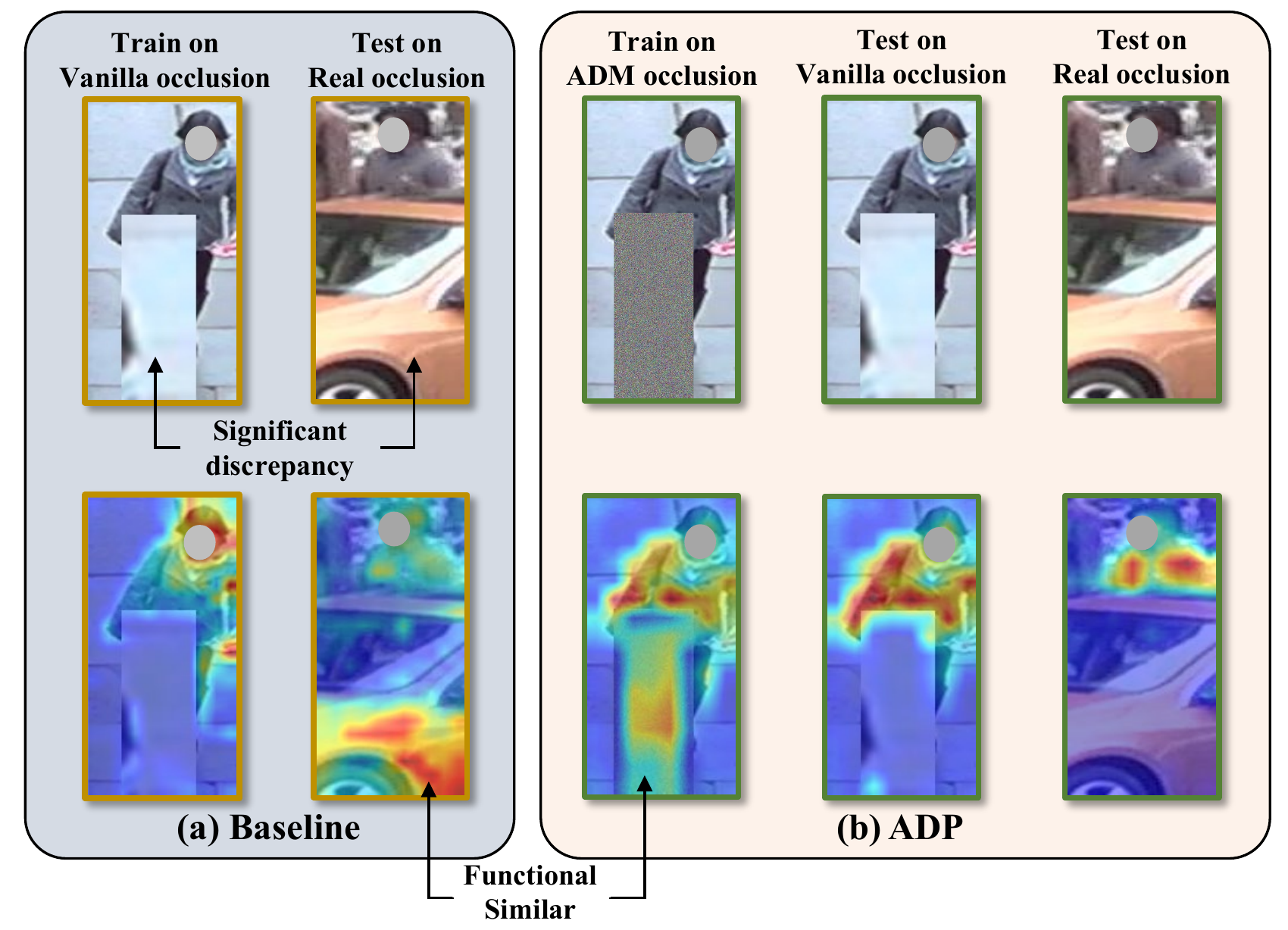}

    \caption{Visualization of attention to baseline and proposed ADP. (a) The baseline trained with the assistance of background occlusion failed to avoid the realistic occlusion in the testing set. (b) The ADP trained by the proposed reality-similar occlusion ADM performs well on both artificial and real occlusion.}
   \label{fig:onecol}
\end{figure}
To address the aforementioned problems, some methods~\cite{DBLP:conf/cvpr/WangYLWYWYZS20, DBLP:conf/cvpr/GaoWLL20, DBLP:conf/iccv/MiaoWLD019, DBLP:journals/tnn/MiaoWY22} use additional trained networks, such as human parsing and keypoint estimation, to align different human parts.
With the aid of these extra networks, the occluded parts can be repaired by disseminating information from the visible parts.
However, these approaches are severely limited due to the domain gap between the pre-trained network and the Re-ID dataset.
%

%
Recently, with the exploration of attention mechanisms for various vision tasks, it has also been adopted for occluded person Re-ID to eliminate the interference of noisy information~\cite{DBLP:journals/tip/ZhaoLDZWW21, DBLP:conf/cvpr/SunXLZLWS19, DBLP:conf/iccv/He0WW0021}.
During the process of attention learning, many data augmentation strategies~\cite{DBLP:conf/iccv/ChenLDLYXCJ21, DBLP:conf/cvpr/WangZT0HS22, DBLP:conf/icmcs/ZhuoCLW18} generate artificial occlusion, which directs the attention to person and forces it to avoid occluded regions.
Currently, the most widely used artificial occlusion methods are random erasing~\cite{zhong2020random} or using the background as occlusion~\cite{DBLP:conf/iccv/ChenLDLYXCJ21}.
Nevertheless, pre-trained attention networks are inherently more likely to focus on the semantically rich foreground than the background.
Therefore, network will inevitably tend to ignore the occlusion constituted by the background, which will result in a lack of generalization.
%
%
To illustrate this point, we utilized background for artificial occlusion based on the ViT baseline in TransReID~\cite{DBLP:conf/iccv/He0WW0021} and visualize the attention for both the training and testing sets in Fig.\ref{fig:onecol}(a). The results demonstrate that while the baseline can avoid artificial occlusions well in the training set, attention is still disturbed in the testing set due to the significant discrepancy between artificial occlusion and actual occlusion.

In this paper, we propose a solution to the challenges mentioned above by introducing an Attention Disturbance Mask (ADM) module that simulates real-world occlusions with greater fidelity. The primary way in which occlusions disrupt models is by impeding attention. However, obtaining enough occluded data to enable the model to avoid such disruptions is difficult.
To surmount this problem, we utilize an attack-oriented methodology that produces noise masks with the capacity to simulate the interference effects of actual obstructions at the feature level. This enables us to construct occlusions that mirror the effects of those encountered in real-world scenarios. As illustrated in Figure \ref{fig:onecol}(b), the proposed Attention Disturbance Module (ADM) performs a similar role to real-world occlusions by introducing disruptions to the neural network's attention. This finding directly verifies the capability of our designed ADM in faithfully emulating occlusions at the feature level. By training the network on such occlusions that closely resemble those encountered in real-world scenarios, we can effectively enhance its robustness against occlusions during testing.

However, handling complex occlusions directly can pose optimization challenges for the network. To address this issue, we propose the Dual-Path Constraint Module (DPC) to handle both holistic and occluded images simultaneously, thus using holistic features as an extra supervisor to guide attention more towards the target pedestrian. Notably, the network parameters in the proposed DPC are shared by both paths, while the individual classifiers learn information about holistic and occluded images separately.

%
%
%
%
The main contributions of our method can be summarized as below:

\begin{itemize}
    \item
    We first introduce a novel attack-based augmentation strategy called the Attention Disturbance Mask (ADM), which simulates real occlusion at the feature level and effectively diverts attention away from actual occlusions during testing.
    \item
    We propose a Dual-Path Constraint module (DPC) that utilizes dual-path interactions to encourage the network to learn a more generalized attention mechanism. DPC is compatible with existing occlusion-based data augmentation methods and can provide significant performance improvements.
    %
    \item The two proposed methods are both used to assist in the training of the baseline, and can be discarded in the inference stage, making them easy to be compatible with many existing methods, indicating the efficiency and wide applicability of our method.
    \item Trained with our proposed ADP, the transformer baseline can achieve new state-of-the-art performance on multiple benchmark datasets \emph{e.g.}, $74.5\%$ on Rank-1 on Occluded-Duke dataset. 
\end{itemize}
%
\section{Related Work}
%
\begin{figure*}[t]
  \centering
  \includegraphics[width=1\linewidth]{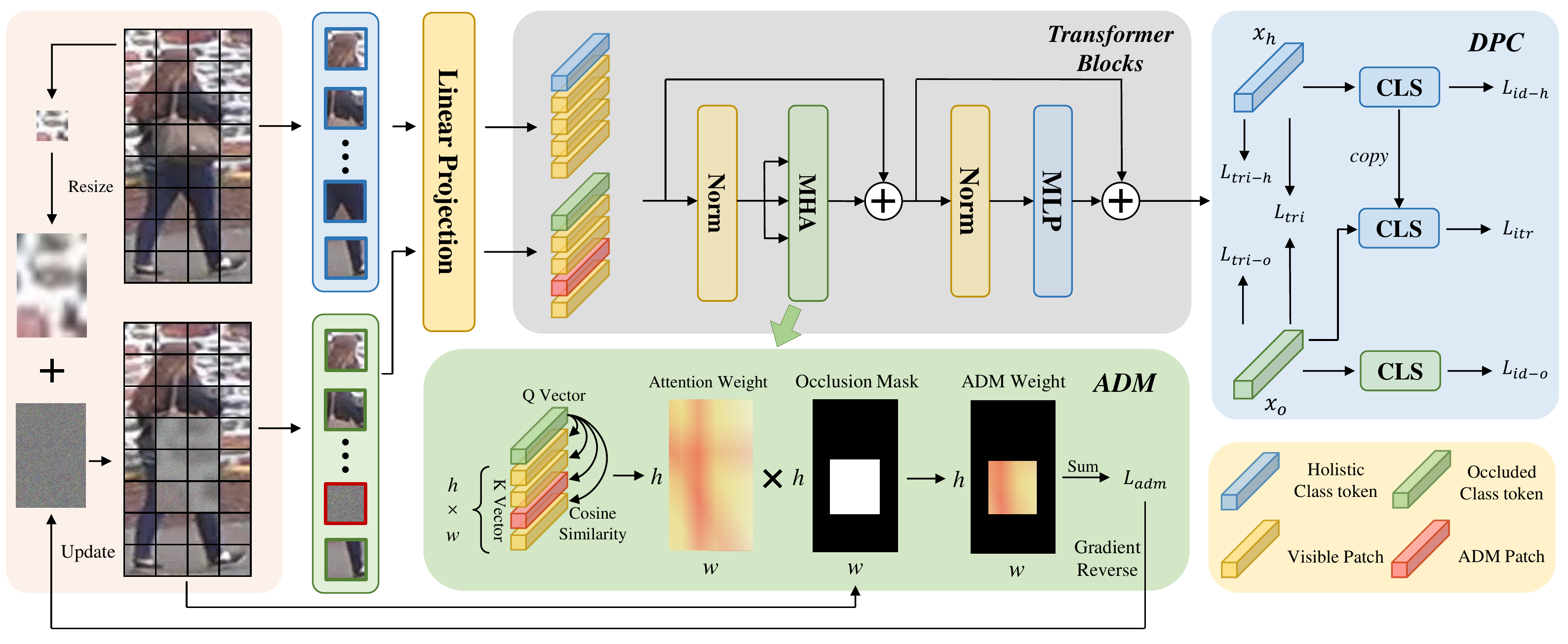}
  \caption{The overview of the proposed Attention Disturbance and Dual-Path Constraint Network (ADP). To create the corresponding occlusion images, the transformed background patch is used as the carrier of the Attention Disturbance Mask (ADM) and covers a random region of the original image. Then the Dual-Path Constraint Module (DPC) simultaneously deals with holistic and occluded images. In Multi-Head Attention (MHA) stage, ADM maximizes the similarity between class token and masked patches to optimize the mask.}
  \label{fig:short}
\end{figure*}
%
%
%
Two main issues of occluded person Re-ID are the missing information and the noisy information caused by the various obstacles.
%
%
Some methods have been proposed to address the missing information issue by attempting to remove the obstacle and reconstruct the missing visible human parts~\cite{DBLP:conf/cvpr/HouMCGSC19,DBLP:journals/pami/HouMCGSC22, DBLP:conf/cvpr/WangYLWYWYZS20}. 
Hou \emph{et al.}~\cite{DBLP:conf/cvpr/HouMCGSC19} designed an auto-encoder to generate contents of the occluded parts at pixel level.
Wang \emph{et al.}~\cite{DBLP:conf/cvpr/WangYLWYWYZS20} utilize an additional pose estimate network to detect the key-points then find the high-order relation and human-topology information.
Moreover, an adaptive direction graph convolutional layer is proposed to pass relation information from visible to occluded nodes.
Hou \emph{et al.}~\cite{DBLP:journals/pami/HouMCGSC22} divide the feature map into six regions and predict the features of occluded regions by adopting the long-range spatial contexts from non-occluded regions.
Instead of reconstructing the missing parts, other methods choose to more focus on the visible parts.
With the help of the attention mechanism, these methods can generate attention maps in which occluded regions are given smaller weights to discard the noisy information.
%
%
To better discover the occluded part, several data augmentation strategies are adopted.
Zhuo \emph{et al.}~\cite{DBLP:conf/icmcs/ZhuoCLW18} design an occlusion simulator to use the random patch from the background as the artificial occlusion to cover the full-body person image.
Chen \emph{et al.}~\cite{DBLP:conf/iccv/ChenLDLYXCJ21} also use the background information and paste it to eight pre-defined locations to make the data augmentation.
%

%
These artificial occlusion data augmentation can provide labels for the location of occlusions. This extra information helps train attention mechanisms to exclude noisy occlusions. However, networks trained on artificial occlusions often cannot handle realistic occlusions well, due to the significant discrepancy between artificial and realistic occlusion.
The proposed ADP method takes a different approach. It generates more realistic occlusions while also providing extra supervision from holistic features. The more realistic occlusions lead to more robust and generalized networks.
\section{Proposed Method}
\subsection{Overview}
%
In this section, we introduce the proposed Attention Disturbance and Dual-Path Constraint Network (ADP). The architecture of ADP is depicted in Fig.\ref{fig:short}, where a pre-trained ViT~\cite{DBLP:conf/iclr/DosovitskiyB0WZ21} is utilized as the backbone to extract image features. To generate occluded images, we use an artificial occlusion consisting of a background patch and an attention disturbance mask. Both holistic and occluded images are fed into a parameter-shared transformer to extract their respective features. Similar to TransReID~\cite{DBLP:conf/iccv/He0WW0021}, we divide the input images, which have a resolution of H × W, into N patches and then convert them into patch embeddings via a linear projection operation with a patch size of P and a stride size of S.
\begin{equation}
N= h \times w =\left\lfloor\frac{H+S-P}{S}\right\rfloor \times\left\lfloor\frac{W+S-P}{S}\right\rfloor.
\end{equation} 
Meanwhile, a learnable class embedding token denoted as $x_{cls}$ is attached to the input sequences to aggregate the image information during the process and act as the global feature map in the output stage.
Besides, a learnable position embedding $\mathcal{P} \in \mathbb{R}^{(N+1) \times d}$ is added to the transformer to append spatial information to the transformer.
The complete input sequence embeddings can be formulated as:
\begin{equation}
\mathcal{Z}_{0}=\left[x_{\mathrm{cls}} ; \mathcal{F}\left(x_{p}^{1}\right) ; \mathcal{F}\left(x_{p}^{2}\right) ; \cdots ; \mathcal{F}\left(x_{p}^{N}\right)\right]+\mathcal{P},
\end{equation} 
where $\mathcal{Z}_0$ represents input sequence embeddings, $\mathcal{F}$ is the linear projection operation mapping the input image $x \in \mathbb{R}^{H \times W \times C}$ into patch embedding $ f_{pe} \in \mathbb{R}^{N \times d}$, $N$ is the number of flattened $h \times w$ patched, and $d$ denotes the dimensions.
Then, the patch embedding $ f_{pe}$ is combined with the class token as the input feature map of transformer blocks.
\subsection{Attention Disturbance Mask}
The main idea of ADM is to generate a mask that can make the network's attention accidentally focus on the occlusion, which simulates the same effect as real-world occlusion.
However, directly generating a mask to make attention focus on a blank area is difficult.
Therefore, the background information is adopted as a carrier of mask to provide the domain information and simplify the mask optimization process.
To get the background region of the input image, we randomly select the corner patch of image.
Then the cropped background patch ${P}_b$ will be resized to $s_o = r_o \times s$, where $r_o\sim \mathcal{U}(0.1,0.5)$ and $s = H \times W$.
The shape of ${P}_b$ is $H_b = \sqrt{s_o \times r_s}$ and $W_b = \sqrt{\frac{s_o}{r_s} }$ with $r_s \sim \mathcal{U}(0.3, 3.3) $.
The processed background patch is pasted arbitrarily anywhere in the image, and a mask $\mathcal{M} \in \mathbb{R}^{H \times W}$ corresponding to the pasting position is saved for overlay attention disturbance mask.
To generate the ADM, we first initialize a random learnable parameter, which has same shape as input image, and superimpose it to the occlusion position according to the $\mathcal{M}$.
%
During training, we dynamically update the ADM based on the weight matrix in the multi-head attention stage of each transformer block.
Considering the mechanism of attention operation, it first calculates the dot-product between the queries $Q$ and the keys $K$ to measure the similarity between them and in accordance with the similarity to get the weight matrix~\cite{DBLP:conf/nips/VaswaniSPUJGKP17}.
The whole attention process can be formulated as follows:
\begin{equation}
\operatorname{Attention}(Q, K, V) = \mathbf{W}  V,
\end{equation} 
\begin{equation}
\mathbf{W} = softmax(\frac{\mathbf{Q}\mathbf{K}^\mathbf{T}}{\sqrt{{c}/{N_H}}}),
\end{equation} 
%
where $N_H$ means the number of heads in multi-head attention.
Therefore, we can disturb the attention by maximizing the similarity between the class token and the occluded region, which will increase the occluded region's weight and make attention mistakenly focus on it.
To be specific, in each transformer block, we have class token embedding $x_{cls}^i$ and patch embedding $x_p^i$, where $i$ represents the $i$-th block of the transformer.
Then the disturbance loss can be formulated as:
\begin{equation}
\mathcal{L}_{adm} = \sum_{i=1}^N \sum_n softmax(\frac{\mathbf{x_{cls}^i}\mathbf{x_p^i}^\mathbf{T}}{\sqrt{{c}/{N_H}}}) \odot \tilde{\mathcal{M}},
\end{equation} 
$\tilde{\mathcal{M}}$ represents the resized occlusion mask $\mathcal{M}$ according to the patch size, $\odot$ presents element-wise product.
Then we adopt an additional optimizer to update the ADM alone based on the reversed gradient of the disturbance loss.

\subsection{Dual-Path Constraint Module}
By generating a corresponding occluded image for each input image, we can obtain the holistic-occluded paired data and make it possible to exploit the holistic image as additional supervision for occluded images, which will help deal with complex occlusion cases.
We separately use the identity loss and metric loss in both holistic and occluded paths to ensure the extracted features towards respective images are reliable and discriminative.
Meanwhile, a global metric loss and information passing classifier is adopted to convey information between two paths.
After extracting the features of holistic and occlusion images, we obtain the final class token of each path as the feature map denoted as $x_h$ and $x_o$.
In the holistic path, we use the cross-entropy loss as ID-loss and triplet loss as metric loss.
The id-loss $\mathcal{L}_{id-h}$ and metric loss $\mathcal{L}_{tri-h}$ are shown as follows:
\begin{equation}
\mathcal{L}_{id-h}=-\frac{1}{B} \sum_{i=1}^{B} \log \frac{e^{(W_{h}^{y_{i}})^T x_{h}^{i}}}{\sum_{j=1}^{C}e^{(W_{h}^{y_j})^T x_{h}^{i}}},
\end{equation} 
\begin{equation}
\mathcal{L}_{tri-h}=\left[ \alpha +  \left(\left\|f_{a}-f_{p}\right\|_{2}^{2}-\left\|f_{a}-f_{n}\right\|_{2}^{2}\right)\right]_+,
\end{equation} 
where the B and C in ID-loss refer to the batch size and the number of class, and $W_{h}$ represents the weight of holistic classifier.
The $f_{a},f_{p}$, and $f_{n}$ in triplet loss refer to the anchor, positive and negative features with online hard-mining~\cite{DBLP:conf/cvpr/SchroffKP15}, and $\alpha$ is the margin.
The loss of holistic path can be calculated as:
\begin{equation}
\mathcal{L}_{h} = \mathcal{L}_{id-h} + \mathcal{L}_{tri-h}.
\end{equation} 
In the occluded path, the existence of occlusion weakens the identity information and fuzzes the inter-class discrepancy when they have the same occlusion.
The widely used softmax loss is incapable of achieving splendid enough intra-class compactness in such difficult conditions.
So we adopt an extra angular margin in the original softmax loss to increase intra-class compactness and inter-class discrepancy~\cite{DBLP:conf/cvpr/DengGXZ19, DBLP:conf/mm/TanDJW22}.
The id-loss $\mathcal{L}_{id-o}$ with the extra margin can be represented as:
\begin{equation}
\begin{split}
\mathcal{L}_{id-o}=-&\frac{1}{B} \sum_{i=1}^{B} \log \frac{e^{s\left(\theta_i +m\right)}}{e^{s\left(\theta_i+m\right)}+\sum_{j=1, j \neq y_{i}}^{C} e^{s(\theta_j)}},
\\
&\theta_i = (W_{o}^{y_{i}})^T x_{o}^{i}, \ \ \theta_j = (W_{o}^{y_{j}})^T x_{o}^{j} \label{con:ido},
\end{split}
\end{equation} 
where $m$ denotes the angular margin and $s$ is the scale adjust hyper-parameter.
And metric loss $\mathcal{L}_{tri-o}$ is also adopted, which is same as holistic path.
The loss of occluded path can be calculated as:
\begin{equation}
\mathcal{L}_{o} = \mathcal{L}_{id-o} + \mathcal{L}_{tri-o}.
\end{equation}
To connect the holistic and occluded path, we adopt a global triplet loss $\mathcal{L}_{tri}$ to close the distance between these two paths.
Furthermore, since the classifier can be regarded as a prototype center of each identity, we use it as an anchor of the holistic identity feature to pull close the same identity in the occluded path to mitigate the gap between them. 
And benefits from the asymmetric structure of the classifier in two paths, the interaction between two paths will bring more information and provide more substantial supervision.
Specifically, we clone the parameter of the classifier in holistic path as $\hat{W}_h$ and calculate the similarity with occluded feature as the interaction loss to eliminate the effect of occlusion.
The interaction loss $\mathcal{L}_{itr}$ can be given as:
\begin{equation}
\mathcal{L}_{itr}=-\frac{1}{B} \sum_{i=1}^{B} \log \frac{e^{(\hat{W}_h^{y_{i}})^T x_{o}^{i}}}{\sum_{j=1}^{C} e^{(\hat{W}_h^{y_j})^T x_{o}^{i}}}.
\end{equation} 
The whole loss function of Dual-Path Constraint Module can be summarized as:
\begin{equation}
\mathcal{L}_{dpc} = \mathcal{L}_{h} + \mathcal{L}_{o} + \mathcal{L}_{tri} + \lambda\mathcal{L}_{itr}, \label{con:dpc}
\end{equation} 
where the $\lambda$ is the hyper-parameter coefficient of $\mathcal{L}_{itr}$.

%

%
In the testing phase, only the pure ViT baseline is used to extract feature maps without any artificial occlusion, which makes our network simple and efficient for implementation.
\begin{table}
\centering
\resizebox{0.48\textwidth}{!}
{
\begin{tabular}{l|cc|cc}
\hline
\multicolumn{1}{c|}{\multirow{2}{*}{Methods}} & \multicolumn{2}{c|}{Occ-Duke} & \multicolumn{2}{c}{Occ-REID} \\ \cline{2-5} 
\multicolumn{1}{c|}{}                         & R-1           & \emph{m}AP             & R-1          & \emph{m}AP             \\ \hline
PCB\cite{DBLP:conf/eccv/SunZYTW18}            & 42.6             & 33.7            & 41.3            & 38.9            \\
DSR\cite{DBLP:conf/cvpr/HeLLS18}              & 40.8             & 30.4            & 72.8            & 62.8            \\
FPR\cite{DBLP:conf/iccv/HeWLZSF19}            & -                & -               & 78.3            & 68.0            \\
Ad-Occ\cite{DBLP:conf/cvpr/HuangL0CH18}  & 44.5             & 32.2            & -               & -               \\
PVPM\cite{DBLP:conf/cvpr/GaoWLL20}            & 47.0             & 37.7            & 66.8            & 59.5            \\
GASM\cite{DBLP:conf/eccv/HeL20}             & -                & -               & 74.5            & 65.6            \\
HOReID\cite{DBLP:conf/cvpr/WangYLWYWYZS20}    & 55.1             & 43.8            & 80.3            & 70.2            \\
OAMN\cite{DBLP:conf/iccv/ChenLDLYXCJ21}       & 62.6             & 46.1            & -               & -               \\
Part-Label\cite{DBLP:conf/iccv/YangZYJZSCZ21} & 62.2             & 46.3            & 81.0            & 71.0            \\ 
ISP\cite{DBLP:conf/eccv/ZhuGLTW20}            & 62.8             & 52.3            & -               & -               \\
PRE-Net\cite{DBLP:journals/tcsv/YanWGYG23}    & 68.3             & 55.2            & -               & -            \\
CAAO\cite{DBLP:journals/tip/ZhaoQJTB23}       & 68.5             & 59.5            & 87.1            & 83.4            \\\hline
PAT\cite{DBLP:conf/cvpr/LiHZL0021}            & 64.5             & 53.6            & 81.6            & 72.1            \\
TransReID\cite{DBLP:conf/iccv/He0WW0021}      & 64.2             & 55.7            & -               & -               \\
PFD\cite{DBLP:conf/aaai/WangLS0S22}           & 67.7             & 60.1            & 79.8            & 81.3            \\
FED\cite{DBLP:conf/cvpr/WangZT0HS22}          & 68.1             & 56.4            & 86.3            & 79.3            \\ 
SAP\cite{DBLP:conf/aaai/Jia0ZCYL23}           & 70.0             & 62.2            & 83.0            & 76.8            \\ \hline
\textbf{ADP(Ours)}                            & 72.2    & 60.6   & 88.2   & 82.0   \\ \hline
TransReID$^\ast$\cite{DBLP:conf/iccv/He0WW0021}& 66.4             & 59.2            & -               & -               \\
PFD$^\ast$\cite{DBLP:conf/aaai/WangLS0S22}    & 69.5             & 61.8            & 81.5            & 83.0            \\
DPM$^\ast$\cite{DBLP:conf/mm/TanDJW22}        & 71.4             & 61.8            & 85.5            & 79.7            \\ \hline
\textbf{ADP(Ours)}$^\ast$                     & \textbf{74.5}    & \textbf{63.8}   & \textbf{89.2}   & \textbf{85.1}   \\ \hline
\end{tabular}
}
\caption{Comparison with state-of-the-art methods on Occluded-Duke and Occluded-REID. $\ast$ indicates that the backbone has a sliding-window setting and a smaller stride.}
\label{label1}
\end{table}

\section{Experiment}
\subsection{Datasets and Evaluation Setting}
To validate the effectiveness of our proposed method, we perform extensive experiments on publicly available Re-ID datasets, including both occluded~\cite{DBLP:conf/iccv/MiaoWLD019, DBLP:conf/icmcs/ZhuoCLW18} and holistic~\cite{DBLP:conf/iccv/ZhengSTWWT15, DBLP:conf/iccv/ZhengZY17, DBLP:conf/eccv/RistaniSZCT16} datasets.
\noindent
\textbf{Occluded-Duke}~\cite{DBLP:conf/iccv/MiaoWLD019} is a large-scale dataset selected from the DukeMTMC for occluded person re-identification. It consists of 15,618 training images of 702 people, while the query and gallery sets contain 2,210 testing images of 519 people and 17,661 images of 1,110 persons, respectively. Until now, Occluded-Duke is still the most challenging dataset for occluded Re-ID due to the scale of occlusion. 
\noindent
\textbf{Occluded-REID}~\cite{DBLP:conf/icmcs/ZhuoCLW18} is an occluded person dataset captured by mobile cameras. A total of 2,000 images were captured from 200 individuals, each consisting of five full-body images and five occluded images. Following the evaluation protocol of previous works~\cite{DBLP:conf/cvpr/GaoWLL20,DBLP:conf/cvpr/WangYLWYWYZS20}, we trained the model under the training set of Market-1501~\cite{DBLP:conf/iccv/ZhengSTWWT15},while Occluded-REID is used only as a test set.
\noindent
\textbf{Market-1501}~\cite{DBLP:conf/iccv/ZhengSTWWT15} is a widely-used holistic Re-ID dataset captured from 6 cameras. The training set contains 1,236 images of 751 people, while the query and gallery sets contain 3,368 images of 750 people and 19,732 images of 750 people, respectively.
\noindent
\textbf{DukeMTMC-reID}~\cite{DBLP:conf/iccv/ZhengZY17, DBLP:conf/eccv/RistaniSZCT16} contains 36,441 images of 1,812 persons captured by 8 cameras, with 16,522 images of 702 identities are used as the training set and 2,228 and 16,522 images of 702 people who do not appear in the training set as the query and gallery images, respectively.

\noindent
\textbf{Evaluation Protocol.} To make it fair compared with other methods, we adopt the widely-used Cumulative Matching Characteristic (CMC) and mean Average Precision (mAP) as evaluation metrics. All experiments are performed in the single query setting.
\noindent
\textbf{Implementation details.} We adopt the ViT~\cite{DBLP:conf/iclr/DosovitskiyB0WZ21} pre-trained on ImageNet~\cite{DBLP:conf/cvpr/DengDSLL009} as our backbone and use 12 transformer blocks with 8 heads for multi-head attention. The numbers of channel is set to 768.
The input images are resized to 256 $\times$ 128 and augmented by commonly used random horizontal flipping, padding and random cropping.
During the training phase, the batch size is set to 64 with 16 identities. We utilize the SGD as the optimizer, with the initial learning rate of 0.004 and a cosine learning rate decay.
The margin of each triplet loss is set to 0.3.
The hyper-parameter $m$ and $s$ in eq.(\ref{con:ido}) are set to 0.3 and 30, respectively, while the $\lambda$ in eq.(\ref{con:dpc}) is 0.1.

\begin{table}
\centering
\resizebox{0.48\textwidth}{!}
{
\begin{tabular}{l|cc|cc}
\hline
\multicolumn{1}{c|}{\multirow{2}{*}{Methods}} & \multicolumn{2}{c|}{Market-1501} & \multicolumn{2}{c}{DukeMTMC} \\ \cline{2-5} 
\multicolumn{1}{c|}{}              & R-1  & \emph{m}AP     & R-1        & \emph{m}AP          \\ \hline
PCB\cite{DBLP:conf/eccv/SunZYTW18} & 92.3    & 77.4    & 81.8          & 66.1         \\
ISP\cite{DBLP:conf/eccv/ZhuGLTW20} & 95.3    & 88.6    & 89.6          & 80.0         \\
BOT\cite{DBLP:conf/cvpr/0004GLL019}& 94.1    & 85.7    & 86.4          & 76.4         \\ \hline
DSR\cite{DBLP:conf/cvpr/HeLLS18}   & 50.7    & 70.0    & 58.8          & 67.2         \\
STNReID\cite{DBLP:journals/tmm/LuoJFZ20}& 66.7 & 80.3    & 54.6          & 71.3         \\
VPM\cite{DBLP:conf/cvpr/SunXLZLWS19}  & 93.0  & 80.8   & 83.6          & 72.6         \\ \hline
HOReID\cite{DBLP:conf/cvpr/WangYLWYWYZS20} & 94.2 & 84.9& 86.9         & 75.6         \\
OAMN\cite{DBLP:conf/iccv/ChenLDLYXCJ21}   & 93.2& 79.8 & 86.3          & 72.6         \\
FPR\cite{DBLP:conf/iccv/HeWLZSF19} & 95.4     & 86.6  & 88.6          & 78.4         \\
PAT\cite{DBLP:conf/cvpr/LiHZL0021}  & 95.4    & 88.0    & 88.8         & 78.2         \\
FED\cite{DBLP:conf/cvpr/WangZT0HS22} & 95.0   & 86.3   & 89.4          & 78.0         \\
TransReID$\ast$\cite{DBLP:conf/iccv/He0WW0021} & 95.2  & 88.9  & 90.7  & 82.0         \\
PRE-Net\cite{DBLP:journals/tcsv/YanWGYG23}  & 95.3     & 86.5    & 89.3 & 77.8            \\
CAAO\cite{DBLP:journals/tip/ZhaoQJTB23}    & 95.3      & 88.0    & 89.8  & 80.9 \\
DPM$\ast$\cite{DBLP:conf/mm/TanDJW22}   & 95.5  & 89.7  & 91.0   & 82.6      \\ 
PFD$\ast$\cite{DBLP:conf/aaai/WangLS0S22}& 95.5& 89.7& \textbf{91.2} & \textbf{83.2} \\
SAP\cite{DBLP:conf/aaai/Jia0ZCYL23}      & \textbf{96.0}     & \textbf{90.5}     & -   & -\\\hline

\textbf{ADP(Ours)$\ast$}    & 95.6   & 89.5 & \textbf{91.2}   & 83.1    \\ \hline
\end{tabular}
}
\caption{Comparison with state-of-the-art methods on Market-1501 and DukeMTMC-reID. $\ast$ indicates that the backbone has a sliding-window setting and a smaller stride.}
\label{label2}
\end{table}

\begin{figure*}[t]
  \centering
  \includegraphics[width=1\linewidth]{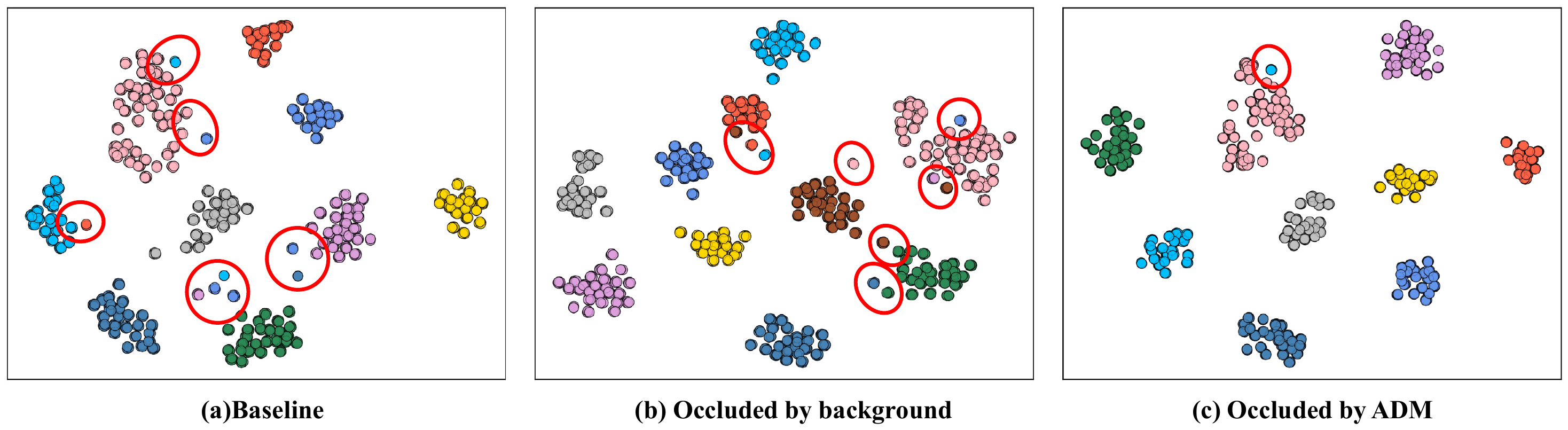}
  \caption{Visualization of the feature distribution on Occlude-Duke dataset. Circles denote the features of images while the colors represent different identities. (a) Baseline refers to the model trained on the images without extra occlusions. (b) The middle plot shows the results of model trained on the images occluded by the background. (c) Compared to the other two models, the model trained on our ADM can avoid the influence of obstacles well.}
  \label{fig:tsne}
\end{figure*}

\subsection{Comparison with State-of-the-art Methods}
\noindent
\textbf{Result on Occluded Datasets.} We compare our ADP with existing state-of-the-art (SOTA) methods on two occluded datasets, and the results are shown in Table~\ref{label1}.
The comparison methods can be divided into CNN-based methods and Transformer-based methods.
%
As can be seen from Table \ref{label1}, transformer-based methods outperform the CNN-based methods. This improvement can be achieved by approximately 15\%, which demonstrates the utilization of attention mechanism is beneficial to occlusion tasks.  
A case in point is that in the most challenging Occluded-Duke dataset, our proposed method ADP can achieve $72.2\%$ in rank-1.
Furthermore, with a small step sliding-window setting, the proposed ADP$^\ast$ can further achieve a higher performance of $74.5\%$ in rank-1 and $63.8\%$ in \emph{m}AP, respectively, exceeds +$3.1\%$ in Rank-1 and +$2.0\%$ in \emph{m}AP compared with the transformer-based SOTA method DPM~\cite{DBLP:conf/mm/TanDJW22}.
%
%
On the Occluded-REID dataset, our ADP and ADP$^\ast$ also consistently outperform current SOTAs.
%
%

\begin{table}
\centering
\resizebox{0.4\textwidth}{!}
{
\begin{tabular}{l|llll}
\hline
\multicolumn{1}{c|}{\multirow{2}{*}{Method}} & \multicolumn{4}{c}{Occluded-Duke}                             \\ \cline{2-5} 
\multicolumn{1}{c|}{}                        & R-1           & R-5           & R-10          & \emph{m}AP           \\ \hline
baseline                                     & 59.7          & 75.3          & 80.7          & 49.8          \\
+ADM                                         & 66.2          & 81.6          & 86.3          & 57.7          \\
+DPC                                     & 72.2          & 85.1          & 88.0          & 60.6          \\  \hline
baseline$^\ast$                              & 63.2          & 78.8          & 83.6          & 53.3          \\
+ADM                                         & 69.8          & 82.8          & 87.5          & 60.3          \\
+DPC                                     & \textbf{74.5} & \textbf{86.4} & \textbf{89.6} & \textbf{63.8} \\ \hline
\end{tabular}
}
\caption{Ablation study of each proposed module in ADP on Occluded-Duke dataset. $\ast$ indicates that the backbone has a sliding-window setting and a smaller stride.}
\label{label3}
\end{table}


%
\noindent
\textbf{Result on Holistic Datasets.} We also experiment our proposed method on holistic person Re-ID datasets, including Market-1501 and DukeMTMC-reID, and compare our method with state-of-the-art methods in three categories, \emph{i.e.}, holistic Re-ID methods~\cite{DBLP:conf/eccv/ZhuGLTW20,DBLP:conf/cvpr/0004GLL019, DBLP:conf/eccv/SunZYTW18}, partial Re-ID methods~\cite{DBLP:journals/tmm/LuoJFZ20, DBLP:conf/cvpr/SunXLZLWS19} and occluded Re-ID methods~\cite{DBLP:conf/cvpr/WangYLWYWYZS20,DBLP:conf/iccv/ChenLDLYXCJ21,DBLP:conf/iccv/HeWLZSF19,DBLP:conf/cvpr/LiHZL0021,DBLP:conf/cvpr/WangZT0HS22,DBLP:conf/iccv/He0WW0021,DBLP:conf/mm/TanDJW22,DBLP:conf/aaai/WangLS0S22, DBLP:conf/aaai/Jia0ZCYL23, DBLP:journals/tcsv/YanWGYG23, DBLP:journals/tip/ZhaoQJTB23} methods.
The results are shown in Table~\ref{label2}.
%
Though designed for occlusion problems, our proposed module achieves comparable performance on holistic datasets.
For example, the proposed ADP can achieve +$0.3\%$/+$1.6\%$ improvement in Rank-1 and +$0.9\%$/+$3.1\%$ in \emph{m}AP on Market-1501 and DukeMTMC-ReID datasets, respectively, compared with the state-of-the-art method ISP~\cite{DBLP:conf/eccv/ZhuGLTW20}.
Our ADP also got +$2.6\%$/+$7.6\%$ improvement in Rank-1 and +$8.7\%$/+$10.5\%$ in \emph{m}AP compared with SOTA partial Re-ID method VPM~\cite{DBLP:conf/cvpr/SunXLZLWS19}.
%
%
\subsection{Ablation Studies}
In this section, we implement the ablation studies based on the Occluded-Duke dataset to analyze the influence of each module of the proposed ADP method.
In our study, the baseline method adopts ViT as the backbone of network, which is trained based on the original softmax loss and triplet loss without any artificial occlusion.
The results of ablation studies are given in Table~\ref{label3}.
From the result, we can observe that training with images occluded by the ADM can significantly improve the model performance, in a way that the performance can be increased by +$6.5\%$ in Rank-1 and +$7.9\%$ in \emph{m}AP, respectively, over baseline. Meanwhile, this improvement of performance can reach +$6.6\%$ in Rank-1 and +$7.0\%$ in \emph{m}AP, respectively, over baseline$^\ast$ which only has a smaller stride.
Besides, with the assistance of DPC, the performance of the model can further increase from $66.2\%$ to $72.2\%$ in Rank-1 and $57.7\%$ to $60.6\%$ in \emph{m}AP over baseline, and increase from $69.8\%$ to $74.5\%$ in Rank-1 and $60.3\%$ to $63.8\%$ in \emph{m}AP over baseline$^\ast$.

\begin{table}
\centering
\resizebox{0.42\textwidth}{!}
{
\begin{tabular}{l|llll}
\hline
\multicolumn{1}{c|}{\multirow{2}{*}{Method}} & \multicolumn{4}{c}{Occluded-Duke}                             \\ \cline{2-5} 
\multicolumn{1}{c|}{}                & R-1           & R-5           & R-10          & \emph{m}AP           \\ \hline
baseline                                     & 59.7          & 75.3          & 80.7          & 49.8          \\
+ADM                                  & 66.2          & 81.6          & 86.3          & 57.7          \\
+AM                                  & 69.5          & 82.2          & 85.9          & 58.8          \\
+DP                                  & 70.5          & 83.4          & 87.0          & 59.3          \\
+${L}_{itr}$                         & 71.8          & 83.3          & 87.1          & 59.7          \\
+${L}_{tri}$(full)                   & \textbf{72.2} & \textbf{85.1} & \textbf{88.0} & \textbf{60.6} \\
\hline
\end{tabular}
}
\caption{Ablation study of the dual-path loss used in DPC on Occluded-Duke dataset. AM denotes the module use shared angular softmax with single-path structure, while DP represents the module with an asymmetric dual-path structure.}
\label{label4}
\end{table}

We next conduct the ablation test to evaluate the influence of structure and loss function in the DPC module on Occluded-Duke dataset.
The result is given in Table~\ref{label4}, which shows the effectiveness of the proposed dual-path structure and adopted loss used for connecting two paths.
Specifically, with the single-path structure, the adopted angular margin softmax does improve the performance, but it is not suitable for dealing with holistic images and is prone to overfitting problem.
To the contrast, since we propose the dual-path structure and separate the holistic and occluded images to use an asymmetric classification, the performance increased by +$1.0\%$ in Rank-1 and +$0.5\%$ in \emph{m}AP.
Meanwhile, benefiting from the proposed dual-path structure, we can add additional connections between two paths to better leverage the advantages of different types of data.
As a result, ${L}_{itr}$ and ${L}_{tri}$ can improve the performance of rank1 by +$1.3\%$/+$0.4\%$ and \emph{m}AP by +$0.4\%$/+$0.9\%$, respectively.

\subsection{Discussions}
\noindent
\textbf{Effectiveness of ADM occlusion.} To better demonstrate the advantages of ADM occlusion, we compare various occlusion schemes, including random erasing~\cite{zhong2020random} and directly using the background as occlusion~\cite{DBLP:conf/iccv/ChenLDLYXCJ21}. The results are shown in Table~\ref{table5}.
From index-2 to index-4, ADM exhibits prominent performance improvement over conventional occlusion schemes.
In detail, compared with random erasing, the performance is significantly increased by +$5.1\%$ in Rank-1 and +$3.9\%$ in \emph{m}AP, respectively;
compared with the background occlusion, ADM can also achieve +$1.4\%$ performance increase in Rank-1 and +$1.2\%$ in \emph{m}AP, respectively.
Meanwhile, we further visualize the distributions of different features extracted by the model trained with different occlusion strategies.
The simulation results are shown in Fig.\ref{fig:tsne}, where the circles denote the features of images randomly selected from testing set of Occluded-Duke dataset and visualized via t-SNE~\cite{van2008visualizing}. 
In detail, Fig.\ref{fig:tsne}(a) illustrates the distribution of features extracted by the baseline. It is evident that there are numerous outlier features caused by occlusions.
%
In Fig.\ref{fig:tsne}(b), the widely used background occlusion is able to reduce the situation of the outlier features to some extent, but it still cannot completely eliminate them, indicating the model is still affected by the obstacles;
To the contrast, with the model trained by our proposed ADM, the outlier features in Fig.\ref{fig:tsne}(c) almost disappear due to the model's excellent ability to avoid obstacles.
In a nutshell, Fig.\ref{fig:tsne} proves that ADM can help the model reduce the impact of obstacles.

\begin{table}
\centering
\resizebox{0.46\textwidth}{!}
{
\begin{tabular}{c|llll|ll}
\hline
Index                  & RE           & BG & ADM       & DPC & Rank-1        & \emph{m}AP           \\ \hline
\multicolumn{1}{c|}{1} &              &    &           &     & 59.7          & 49.8          \\
\multicolumn{1}{c|}{2} & $\ \checkmark$ &    &           &     & 61.1          & 53.8          \\
\multicolumn{1}{c|}{3} &              & $\ \checkmark$&  &     & 64.8          & 56.5          \\
\multicolumn{1}{c|}{4} &              & & $\ \ \checkmark$ &     & \textbf{66.2} &\textbf{57.7}  \\ \hline
\multicolumn{1}{c|}{5} & $\ \checkmark$ &   && $\ \ \checkmark$   & 66.7        & 56.5          \\
\multicolumn{1}{c|}{6} &  & $\ \checkmark$  & & $\ \ \checkmark$  & 70.4        & 58.4          \\
\multicolumn{1}{c|}{7} & &  & $\ \ \checkmark$ & $\ \ \checkmark$ & \textbf{72.2} & \textbf{60.6} \\ \hline
\end{tabular}
}
\caption{Comparison with previous occlusion strategies. RE indicates the random erasing method, while BG denotes the occlusion strategy using background.}
\label{table5}
\end{table}

\noindent
\textbf{Effectiveness of DPC module.} We evaluated the effectiveness of our proposed DPC with different occlusion strategies, and the simulation results are presented in index-5 to index-7 of Table~\ref{table5}. 
Compared with the result in index-2 to index-4, our experimental results demonstrate that the DPC can be seamlessly integrated into each occlusion strategy and show improvements in performance.
For example, in the random erasing strategy, performance can be increased from $61.1\%$ to $66.7\%$ in Rank-1 and $53.8\%$ to $56.5\%$ in \emph{m}AP, respectively.
Moreover, with the background occlusion, the DPC can significantly increase the Rank-1 performance by +$5.6\%$ and \emph{m}AP performance by +$1.9\%$, respectively.
Our results indicate that the proposed DPC has strong compatibility and universality, making it capable of improving the performance of several previous methods.

\subsection{Visualization of the Attention}
\begin{figure}[t]
  \centering
   \includegraphics[width=1\linewidth]{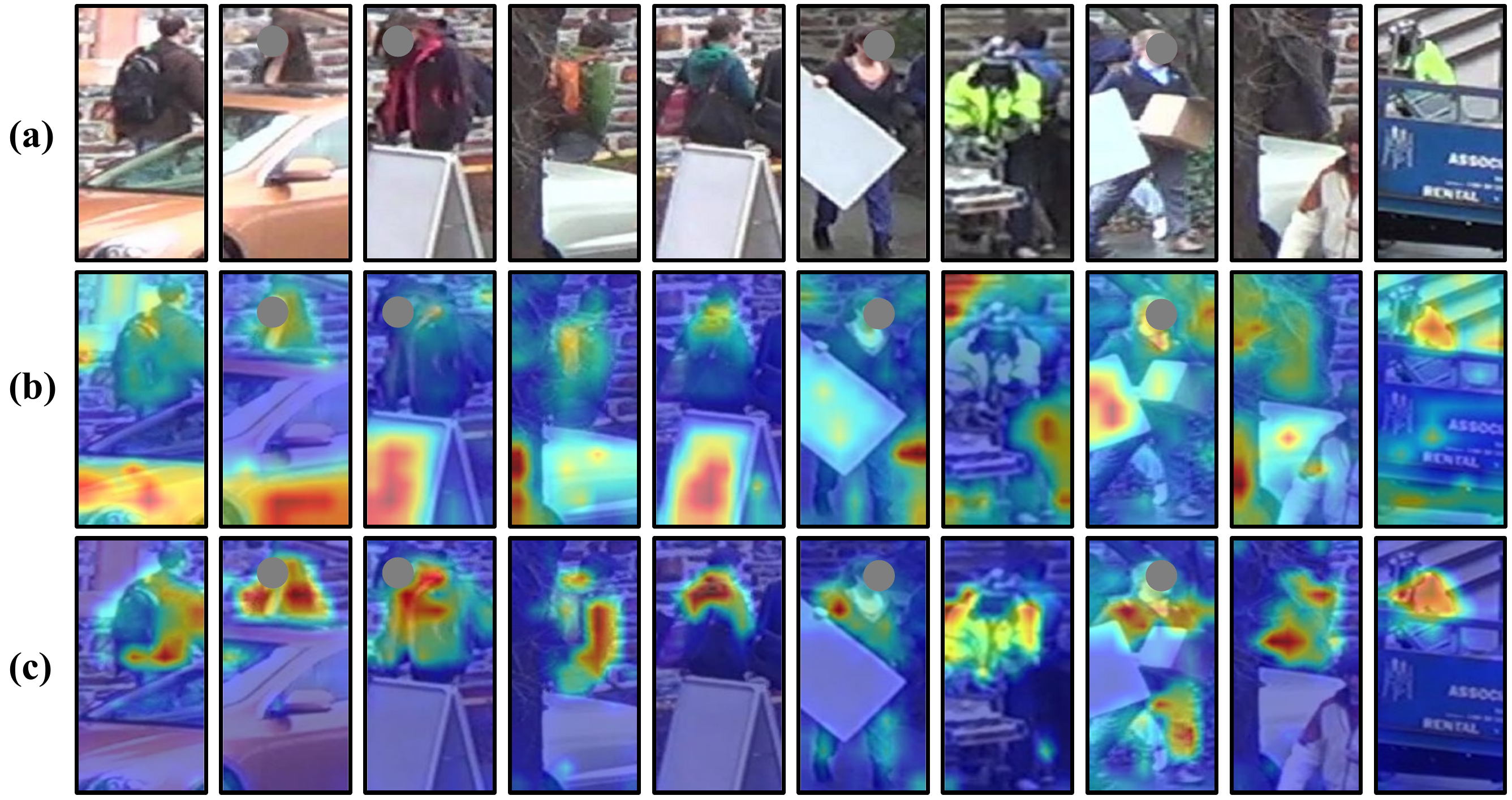}
   \caption{Visualization of attention maps on occlusion testing set in Occluded-Duke dataset. (a) occluded person images. (b) attention maps of baseline. (c) attention maps of ADP.}
   \label{fig:heatmap}
\end{figure}

%
To demonstrate the model's ability to process images with occlusions, we visualize the attention maps and show in Fig.\ref{fig:heatmap}.
The input images are from the testing set with diverse occlusions, and we apply Grad-CAM~\cite{DBLP:conf/iccv/SelvarajuCDVPB17} to visualize the attention heatmap to demonstrate the areas the model focuses on.
It is obvious that baseline can be easily interfered by obstacles, which greatly limits the performance.
In the contrast, Fig.\ref{fig:heatmap}(c) appears that our model's attention mechanism is capable of avoiding paying attention to occlusions to a great extent and focusing more on the target pedestrian.
Furthermore, the attention heatmap shows the proposed model can provide good performance when handling diverse occlusion types and locations.

\section{Conclusion}
%
%

In this research, we introduced a new approach to address the problem of occluded person re-identification by proposing two innovative modules. The ADM generates a more effective artificial occlusion that closely resembles real-world occlusions at the feature level, making the network robust to unseen occlusions and enhancing its generalization. The DPC handles both holistic and occluded images simultaneously, aligning the holistic and occluded features and guiding attention more toward the target pedestrian. Meanwhile, the two proposed modules, ADM and DPC, can be seamlessly integrated with various existing models to enhance their performance, demonstrating the wide applicability of our approach. Experiment results on two occluded datasets and two holistic datasets, illustrate the effectiveness of proposed method and superiority to other state-of-the-art methods.

\section*{Acknowledgements}
This work was supported by National Key R\&D Program of China (No.2022ZD0118202), the National Science Fund for Distinguished Young Scholars (No.620256 03), the National Natural Science Foundation of China (No. U21B2037, No. U22B2051, No. 62176222, No. 62176223, No. 62176226, No. 62072386, No. 620723 87, No. 62072389, No. 62002305 and No. 62272401), and the Natural Science Foundation of Fujian Province of China (No.2021J01002,  No.2022J06001).
\bibliography{aaai24}

\end{document}